\pdfoutput=1
\newcommand{\llama}{Llama}

\documentclass[11pt]{article}

\usepackage[final]{acl}
\usepackage{booktabs}
\usepackage{multirow}
\usepackage{tipa}
\usepackage{times}
\usepackage{latexsym}

\usepackage[T1]{fontenc}

\usepackage[utf8]{inputenc}

\usepackage{microtype}

\usepackage{inconsolata}
\usepackage{multicol}

\usepackage[export]{adjustbox}
\usepackage{graphicx}
\usepackage{listings}

\title{Domain Regeneration: \\
How well do LLMs match syntactic properties of text domains?}

\author{Da Ju \\
Meta AI \\
  \texttt{daju@meta.com} \\\And
  Hagen Blix \\
  New York University \\
  \texttt{hagen.blix@nyu.edu} \\\And
  Adina Williams \\
  FAIR, Meta AI \\
  adinawilliams@meta.com
  }

\begin{document}
\maketitle
\begin{abstract}
Recent improvements in large language model performance have, in all likelihood, been accompanied by improvements in how well they can approximate the distribution of their training data. In this work, we explore the following question: which properties of text domains do LLMs faithfully approximate, and how well do they do so? Applying observational approaches familiar from corpus linguistics, we prompt commonly used, opensource LLMs to regenerate text from three domains of permissively licensed English text which are often contained in LLM training data---Wikipedia, news text, and ELI5. In a fairly semantically-controlled setting, this regeneration paradigm allows us to investigate whether LLMs can faithfully match original human text domains. We investigate varying levels of syntactic abstraction, from simpler properties like sentence length, and article readability, to more complex and higher order properties such as dependency tag distribution, parse depth, and parse complexity. We find that the majority of the regenerated distributions show a shifted mean, a lower standard deviation, and a reduction of the long tail, as compared to the human originals.
\end{abstract}

\section{Introduction}

The question of whether models can transfer capabilities across different domains of texts, or \textbf{domain transfer} has a long history in NLP. Domain considerations have contributed greatly to the establishment of the pretrain-finetune paradigm \citep{devlin-etal-2019-bert,liu-etal-2019-roberta} used for training LLMs \citep{ruder-etal-2019-transfer}. Since the rise in prevalence of LLMs, however, there has been little work explicitly verifying whether state-of-the-art LLMs can actually generate text that matches different, well-described, and well-delineated human-generated text domains. 

One might presume that LLMs should be generally competent in matching text distributions, insofar as one could, from a zoomed out perspective, describe the entire process of pretraining itself as a process of fitting a model to a distribution. However, it is also possible that the increasing number of post-training interventions, such as instruction tuning, automatic preference alignment or other kinds of interventions, could affect this. 

Other model design decisions can also impact a models' ability to match a human-generated distribution. LLMs can suffer from \textbf{model collapse} \citep{dohmatob-etal-2024strong, hamilton-2024-detecting, lanchantin-etal-2025-dpo}, whereby the model has trained on outputs from previous models, which can negatively affect the diversity and quality of generations. Even without being trained on synthetic data, models can fail to match the diversity of human-generated data. LLMs often generate similar data patterns \citep{hupkes-etal-2023-taxonomy}. They can repeat words/tokens \citep{juzek-ward-2024-delve}, use less diverse topics \citep{bache-etal-2013-text, alihosseini-etal-2019-jointly} or both \citep{padmakumar-he-2023-does}.

However, most work investigating the ability of LLMs to match the diversity of human written text does so in the context of an unspecified and uncontrolled ``neutral'' domain. Given that there are many, widely described differences in lexical content, style, syntactic structure etc. across domains \citep{biber-1991-variation, dimarco-hirst-1993-computational, dewdney-etal-2001-form, lee-2002-genres, williams-etal-2018-broad, li-etal-2019-domain}, one might wonder whether models can match human diversity in domains with consistent and well controlled style. 

We explore this question with a paradigm that we call \textbf{LLM-regeneration}:~using the beginning of articles from a well-described domain (e.g. Wikipedia, \citealt{fan-gardent-2022-generating}), we prompt the LLM to complete the article, and then compare the regenerated article to the original. This setting allows us to exert more control over the content of the generations, thus making for a clearer picture of distribution match.

We use this regenerated data to explore the diversity of model outputs in a number of interrelated syntactic measurements. Beyond diversity, we also explore two other \textbf{signatures} of imperfect domain matching in this work: difference in the means of the LLM and human distributions, and a reduced long tail where present in the human distribution. When present, we take each of these three signatures to be evidence that the LLMs we study fail to perfectly match the human text.

While past work has indicated that some LLMs generate more homogeneous syntax than humans \citep{shaib-etal-2024-detection}, such investigations have thus far been restricted to part-of-speech tags, and have only been explored for ``neutral'' domain LLM generations. Here, we combine investigations of text domain with syntactic complexity metrics in an  attempt to delimit a reasonably sized problem space; in addition to text readability and sentence length, we explore more abstract metrics for syntactic complexity, including parse depth, unique dependency tag and constituency label count, and Yngve scores. 

Overall, the contributions of this work are: (i)~We define an experimental setting called LLM-regeneration that enables us to measure how well LLMs match human text with some controls over semantics and domain. (ii) We investigate the extent to which LLMs can match the distribution of text from three well-described human domains. (iii) We illustrate three signatures of domain mismatch---diversity, mean shift, and reduction of the long tail---and show how the three are present for several syntactic complexity metrics, as calculated on generations from opensource models from two model families. These results are important, as they can inform us about whether models can genuinely match text domains. Information about whether LLMs differ in syntactic complexity and variability from human-generated text may additionally be useful for detection of synthetic text, or to guide model improvement. 
\section{Methods}
\subsection{Models}
In this work, we mostly utilize the \llama\ family of models, as they are strongly performing models with open weights. For the majority of our experiments, we utilize \llama-V2 \citep{touvron2023llama} 70B instruction finetuned. We also utilize \llama-V3.3 instruction finetuned \citep{meta-etal-2024-llama} 70B and \llama-V3.1-8B for a subset of our experiments as an additional comparison for the model-specificity of our results. We additionally investigate two models from the Mistral family: Mistral-Small-24B-Intruct\footnote{mistralai/Mistral-Small-24B-Instruct-2501} and Ministral-8B-Instruct\footnote{mistralai/Ministral-8B-Instruct-2410} to verify that our findings hold across model family and size.

\subsection{Data}
We are interested in whether SOTA or near-SOTA LLMs can match properties of text corresponding to its domain, and thus need to select datasets that typify domains. However, some practical complications arose when we embarked on selecting datasets. First of all, we needed to consider which text the model was trained on. In principle, if a model was trained on text from a domain, it should be better at matching the distributional properties of text from that domain. However, the precise nature of the training data for LLMs is generally proprietary information, though it is likely that LLMs were trained on many domains. Clearly, it would not be scalable to investigate all of them (nor to determine where their boundaries lay). 

Additionally, we could, in principle, select a set of data and train an LLM from scratch on it. However, due to budgetary constraints, we would doubtless have to focus on a smaller, and likely less performant model. In that case, if we observed differences between the human and model distributions, those differences could just be due to the model being weak, not to anything interesting about the distributions the model had learned.

Given these considerations, we made the practical decision to focus on Wikipedia, a datasource known to be used in open training datasets \citep{workshop2023bloom176bparameteropenaccessmultilingual, soldaini-etal-2024-dolma} and to be a popular and well-studied data source for many NLP applications in English \citep{wu-weld-2010-open, horn-etal-2014-learning, ni-florian-2016-improving, yang-etal-2018-hotpotqa, dinan-etal-2019-wizard, klang-nugues-2019-docria, ein-dor-etal-2019-financial, dinan-etal-2020-multi, nie-etal-2020-adversarial, calixto-etal-2021-wikipedia, eisenschlos-etal-2021-fool, petroni-etal-2021-kilt, semnani-etal-2023-wikichat}. As additional domains, we also investigate the standard news articles dataset, CCNews\footnote{\href{https://huggingface.co/datasets/vblagoje/cc\_news}{https://huggingface.co/datasets/vblagoje/cc\_news}}, and ``Explain Like I'm 5'' dataset (ELI5; \citealt{fan-etal-2019-eli5}) of simply written questions and answers from an online forum. 

Despite their difference in size (our Wikipedia datasets contains roughly 10x more data than CCNews, and CCNews is larger than ELI5), all datasets are representative of consistent and fairly well circumscribed domains. All three datasets are characterized by internally enforced stylistic standards (e.g., due to editors or moderators), and any competent reader could easily match text to the relevant domain. That is, their differences should allow us to determine whether some of the trends we find for Wikipedia are specific to that data source or more general.

\subsection{Data Processing}\label{subsubsec:cleaning}

\paragraph{Data Cleaning.}
Given that Wikipedia data contains a significant amount of structured text, such as lists, titles, urls and citations, we perform a data cleaning stage using the parsing results we collected to enhance signal quality. First, we removed all sentences with fewer than 3 words or more than 500 words. Second, we eliminated all sentences that contained neither a verb nor an auxiliary verb, as identified by POS tagging, to ensure that our data consists of full English sentences. To verify that these filtering steps didn't drive our main results, we plot data ablation results in Appendix~\ref{appendix:ablation_data}, which show the same trends as our main results.

Note that additional data cleaning is an implicit part of our process. We calculate all metrics based on successful Stanza dependency and constituency parsing. Metrics will not be available if an article is empty in the source data, or contains non-English content. Some metrics, such as the depth score and Yngve score, may encounter errors if the tree parsing is excessively deep. 
In cases where articles pass the parsing stage but have a valid article length of zero (meaning the article is not empty in the beginning, but all sentences have been removed due to prior length and POS filtration), we filter out all depth and Yngve scores of zero. 
For sentence-level metrics, we aggregate results from all successfully parsed sentences to complete the calculations. For metrics aggregated at the article level, we exclude any article if any of its sentences fail in parsing or metric calculation. 

\paragraph{Regenerating the Data with LLMs.}

To generate text from the models that replicates our domains of interest, we adopt a similar approach to that used in \citet{ju-etal-2024-female} to ensure that the article topic and content do not vastly differ. 
We isolate the first $256$ words from a Wikipedia article, and the first $180$ words from a CCNews article respectively (since news articles are shorter on average). For ELI5, we prompt the models only with the title of the thread (e.g., \textit{What is an ETF?}). We then feed these into the model using the prompts in Appendix \ref{appendix:prompts}. We use vLLM \cite{kwon2023efficient} for generation with its default coding configurations, including a temperature of $1.0$, which is considered a ``medium'' temperature. 
The resulting articles will be approximately matched to the original articles in their topic and content. We collect all the articles in each domain, which then serve as our regenerated corpus for downstream analysis. We also performed the two data cleaning steps described above on the model-regenerated data as well.

\subsection{Parsing}\label{subsec:parsing}

We employ the data processing pipeline outlined by \citet{williams-etal-2021-relationships} and used in \citet{ju-etal-2024-female} for our analysis. Our pipeline uses the Stanza tool \cite{qi2020stanza} to process the sentences and generate dependency and constituency parses for later analysis. The pipeline consists of:

1. \textbf{Tokenization \& Sentence Segmentation} 


2.~\textbf{Dependency Parsing:} We use the default parser \citep{chen-manning-2014-fast} to dependency parse all text in our experiments.

3. \textbf{Constituency Parsing:} In addition to dependency parsing, we construct a constituency parse for each sentence, which is subsequently used to compute our metrics.

Rarely, a portion of the parsing pipeline would fail (for example, if the sentence in question was merely a set of hyperlinks in the case of some of the Wikipedia data). In that case, the data point would be excluded from our analysis. Details on exclusions can be found in Appendix~\ref{appendix:ablation_data}.

\subsection{Metrics}\label{subsec:metrics}

\begin{table}[t]
    \small
    \begin{tabular}{llccc}
    \toprule
    \multicolumn{2}{c}{Metric} & \multicolumn{3}{c}{Signature} \\ \cmidrule(r){1-2}\cmidrule(r){3-5}
         Type & Domain & $\mu$ & $\sigma$ & Long Tail\\
    \midrule 
         \multirow{3}{7em}{Flesch-Kincaid} & news & $\approx$ & $\searrow$ & reduced\\ 
         & wiki & $\nearrow$ & $\searrow$ & reduced\\  
         & ELI5 & $\searrow$ & $\searrow$ & reduced\\  
         \cmidrule(r){2-5}%

         \multirow{3}{7em}{Dependency}& news & $\nearrow$ & $\searrow$ & n/a \\ 
         & wiki & $\approx$ & $\searrow$ & n/a\\ 
        & ELI5 & $\nearrow$ & $\searrow$ & n/a\\ 
        \cmidrule(r){2-5}
         
         \multirow{3}{7em}{Depth} & news & $\nearrow$ & $\searrow$ & $\approx$\\ 
         & wiki & $\nearrow$ & $\searrow$ & $\approx$\\ 
                 & ELI5 &  $\approx$ &  $\approx$ &  $\approx$\\
                 \cmidrule(r){2-5}

         \multirow{3}{7em}{Yngve}  & news & $\searrow$ & $\searrow$ & reduced \\ 
         & wiki & $\nearrow$ & $\searrow$ & reduced \\ 
        & ELI5 & $\searrow$ & $\searrow$ & reduced\\ 
         \cmidrule(r){2-5}

         \multirow{3}{7em}{Constituency}  & news & $\nearrow$ & $\searrow$ & reduced \\ 
         & wiki & $\nearrow$ & $\searrow$ & reduced \\ 
        & ELI5 & $\approx$  & $\searrow$ & reduced \\ 
        
         \bottomrule
         
    \end{tabular}
    \caption{Schematic description of \llama's distribution shift for our five investigated metrics relative to the human baseline for all text domains. $\mu$ refers to mean shift ($\nearrow$ refers situations when the mean of the distribution is higher for \llama\ than for the human), $\sigma$ to the standard deviation of the distribution($\searrow$ refers to situations where the distribution is narrower for \llama\ than for the original), and `long tail' to whether a heavy right tail that was present in the human distribution is reduced for the \llama\ distribution (`n/a' marks situations with no long tail in the human distribution of the metric).} 
    \label{tab:schematic}
\end{table}

For the majority of our experiments, we plot the distribution of binned scores so that we can compare the regenerated data to the human data for both domains. 
For all metrics, we observe approximately Gaussian distributions for both the original human data and the regenerated data. We plot the overall normal fit line, as calculated by the defaults in Seaborn \citep{Waskom2021} using \texttt{matplotlib} \citep{Hunter:2007} for easy visual inspection. 

Comparing the human data and the LLM-regenerated data, we isolate three observational signatures of domain shift that recur across metrics: the human and the regenerated distributions can differ in \textbf{mean}, they can differ in \textbf{variance}, and they can differ in the presence of a \textbf{long tail} (a heavy right tail). 
A schematic summarizing our results is in \autoref{tab:schematic}. 

\paragraph{Flesch-Kincaid.} First, we measure the Flesch-Kincaid grade level score, following \citet{flesch1948new}. This score pertains to an article overall, and is a standard metric in the education field and in NLP to estimate the reading level of a piece of text, with higher scores being more difficult. The Flesch-Kincaid score relies on words per sentence and syllables per word to derive an estimate of the ease of reading the text snippet. We calculate the Flesch-Kincaid scores using the \texttt{py-readability-scores} library\footnote{\href{https://github.com/cdimascio/py-readability-metrics/tree/master}{https://github.com/cdimascio/py-readability-metrics/tree/master}}, which relies on the Natural Language Toolkit \citep{bird2009natural}. For each dataset, we take each article and calculate its Flesch-Kincaid grade level score, then we consider the scores for all articles as a distribution.\footnote{We observed that removing sentences and restructuring an article during the data cleaning stage leads to issues with readability score calculation. Therefore, we calculated readability scores without data cleaning for articles exceeding 100 words, as shorter articles lack sufficient content for accurate readability assessment.} 

\subsubsection{Syntactic Metrics} For the other four metrics, we relied on syntactic parses, generated following the procedure we described above in \S\ref{subsec:parsing}. Unlike for the Flesch-Kincaid score, for all syntactic metrics, we calculate the result per sentence.  We consider each syntactic metric as a distribution relative to domain and generation source (human vs. LLM). 

\paragraph{Dependency Tags.}
Dependency tags provide a description of the relation between units in a sentence. For each sentence, we count the number of unique dependency tags.

\paragraph{Parse Depths.}
For each sentence, we count the depth of a constituency parse.

\paragraph{Yngve Scores.} As a first qualification on parse depth, we also explore a measure of left vs. right branching parse trees. We measure each sentence's Yngve score \citep{yngve-1960-model} following \citet{roark-etal-2007-syntactic}, which argues that the Yngve metric is useful for diagnosing cognitive impairments. The score roughly corresponds to the deviation of a parse tree from a completely right-branching tree---it is the average number of left branches on the path from the root node to each leaf.

\paragraph{Constituency Labels.}
As a further qualification of parse depth, we also calculate the number of unique constituency labels in a sentence's constituency parse.

\section{Results}
We generally report the results for \llama-70B models in our figures, but more detailed figures that provide additional data for smaller models and/or models from the Mistral family are available in a number of appendices.

\subsection{Descriptive Results}

To situate our results, we first measure some basic dataset statistics. In \autoref{tab:descriptive_statistics}, we present the statistics for the parse tree depth experiment. Recall that some datasets may slightly differ in size based on the success of metric calculation as described above in \S\ref{subsec:metrics}. Also, note that complexity metrics can be correlated with sequence length \citep{salkar-etal-2022-self}.

Across the board, the regenerated data is similar to the original data in terms of words per sentence. For CCnews and Wikipedia, the regenerated data contains more sentences per article, and hence more words per article than the original data. For ELI5, conversely, the regenerated data contains fewer sentences per article, and hence fewer words per article than the original.
Due to our data cleaning (\S\ref{subsubsec:cleaning}), there is some difference in the number of articles preserved for analysis between generation sources (LLM v. human), with slightly more regenerated articles being analyzed. We plot the distribution over sentence lengths in \autoref{fig:Sentence-Length-full} in Appendix~\ref{app:length}. We observe that the regenerated data shows a shifted mean, a reduction in variability and a reduced long tail, when compared to the original human data for each domain.

\begin{table}
\small
\resizebox{\columnwidth}{!}{%
\begin{tabular}{lrrrrrr}
\toprule
Datasets & Articles & Sentences & Words & S/A & W/S & W/A \\
\midrule
CCNews & 0.6M & 12.6M & 0.3B & 21.8 & 24.6 & 535.0 \\
\llama-2-70B & 0.7M & 19.9M & 0.5B & 28.1 & 25.6 & 718.4 \\
\llama-3.3-70B & 0.7M & 23.9M & 0.7B & 34.0 & 27.7 & 941.4 \\
Mistral-24B & 0.7M & 22.7M & 0.5B & 32.1 & 23.9 & 765.1 \\
Mistral-8B & 0.7M & 22.1M & 0.5B & 31.2 & 24.2 & 756.7 \\
\llama-3.1-8B& 0.7M & 19.8M & 0.5B & 28.7 & 26.5 & 759.2 \\
\midrule
Wikipedia & 6.4M & 114.6M & 2.9B & 17.7 & 24.9 & 441.9 \\
\llama-2-70B & 6.6M & 234.0M & 5.4B & 35.4 & 23.2 & 821.0 \\
Mistral-24B & 6.5M & 307.7M & 7.0B & 47.5 & 22.6 & 1076.1 \\
\midrule
ELI5 & 0.6M & 34.2M & 0.6B & 56.2 & 18.6 & 1047.3 \\
\llama-3.3-70B & 0.6M & 9.3M & 0.2B & 15.2 & 20.4 & 310.6 \\
Mistral-24B  & 0.6M & 6.6M & 0.1B & 10.8 & 18.5 & 200.3 \\
Mistral-8B  & 0.6M & 6.1M & 0.1B & 10.2 & 18.7 & 191.3 \\

\bottomrule
\end{tabular}}
\caption{Descriptive statistics (average) on the datasets used for the parse tree experiments. Eligible sentences meet the following criteria: sentence length falls within 3-500 words, and contains $1\leq$ verb or auxiliary verb. 
}
\label{tab:descriptive_statistics}
\end{table}

\subsection{Flesch-Kincaid Scores}
\begin{figure}[!ht]
    \centering
    \includegraphics[width=1\linewidth]{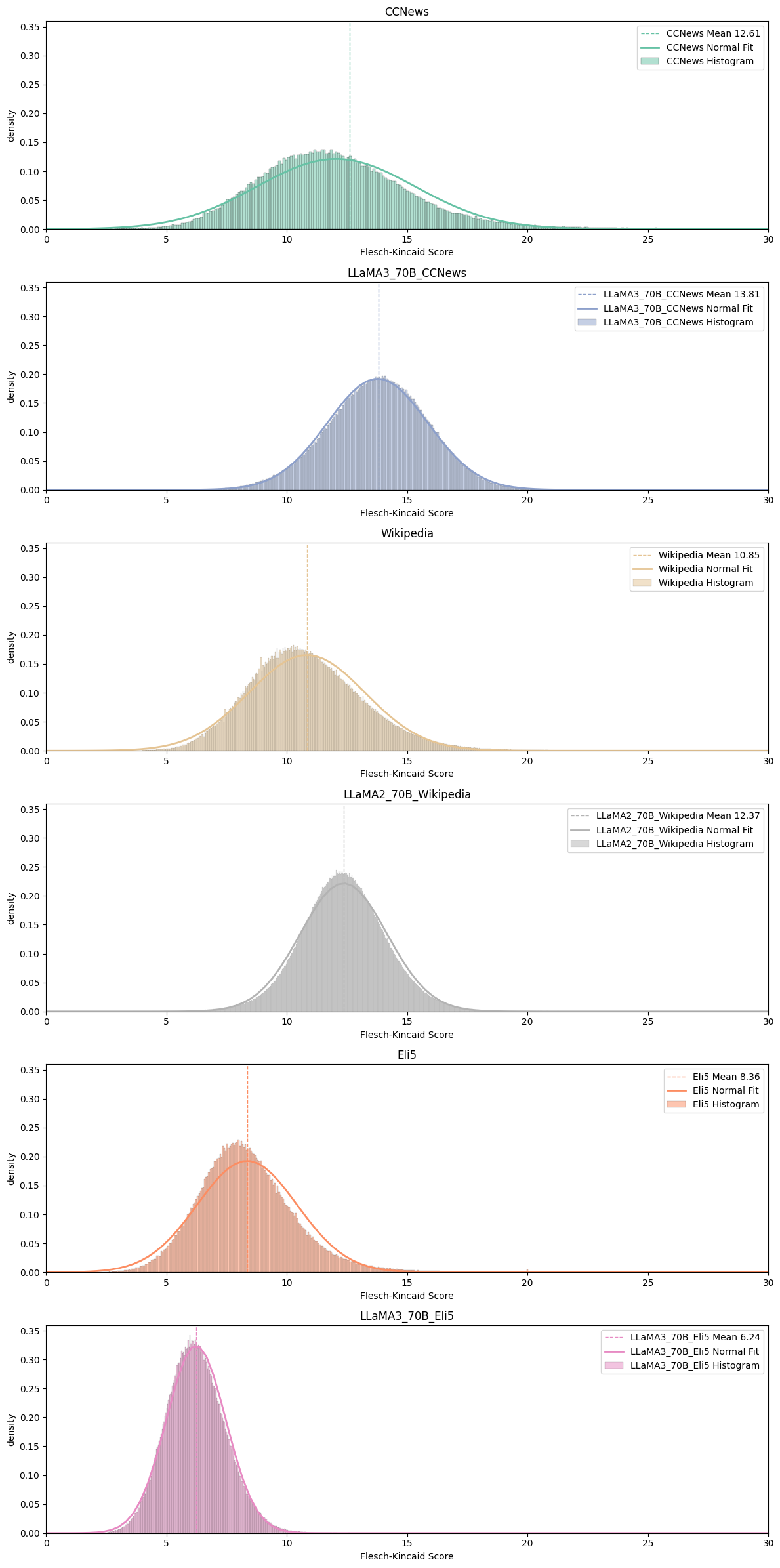}
    \caption{Flesch-Kincaid readability score distributions. Normal fit curves informally illustrate the fact that regenerated data appears to be normally distributed, with narrower distributions and a reduced long tail on the right relative to the human datasets.}
    \label{fig:flesh-kincaid-normal}
\end{figure}

For this simplification metric, shown in Figure~\ref{fig:flesh-kincaid-normal}, we observe that all distributions are roughly normal and the three signatures are present. We observe that Flesch-Kincaid readability scores for the human data deviate slightly from the normal distribution in that they have a right tail, as indicated by the fact that the bars around the center are somewhat above the fit curve on the left, and somewhat below the fit curve on the right. As compared to their human-generated variants, each regenerated dataset has a narrower distribution, and a reduced long tail. For CCNews and Wikipedia, we observe an upward shifted mean, while the mean for ELI5 is shifted downwards.
In Appendix~\ref{appendix:additional_readability}, we plot the full results for all tested models in \autoref{fig:Flesch-Kincaid-full}; we also report the means, medians, and standard deviations for this metric in \autoref{fig:fkgrademeans}, alongside other readability metrics.

\subsection{Dependency Tags}
\begin{figure}[ht]
\includegraphics[width=1\linewidth]{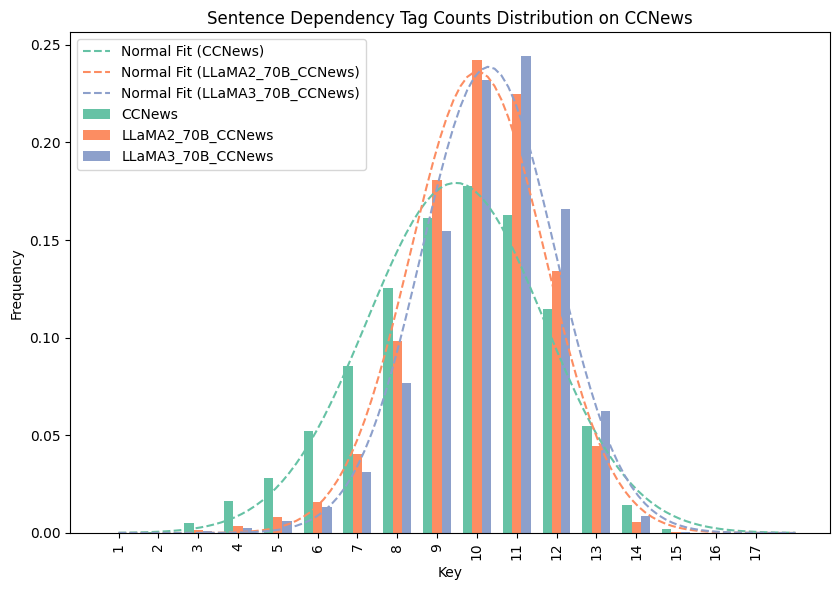}%
   \caption{Proportion of sentences in CCNews that have a particular number of unique dependency tags. 
   }
   \label{fig:deptagccnews}%
   \end{figure}

\begin{figure}[ht]
\includegraphics[width=1\linewidth]{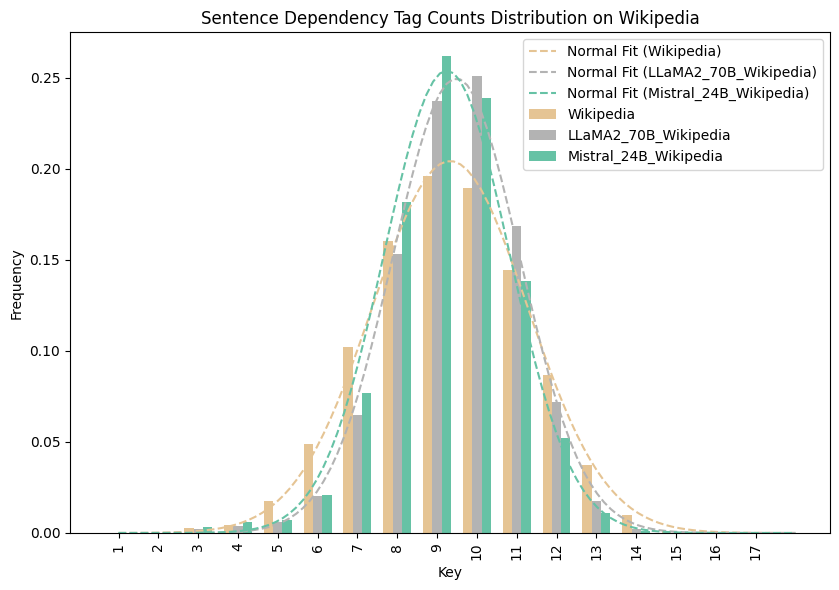}
 \caption{Proportion of sentences in Wikipedia that have a particular number of unique dependency tags. 
 }
\label{fig:deptagwiki}
\end{figure}

\begin{figure}[!h]
\includegraphics[width=1\linewidth]{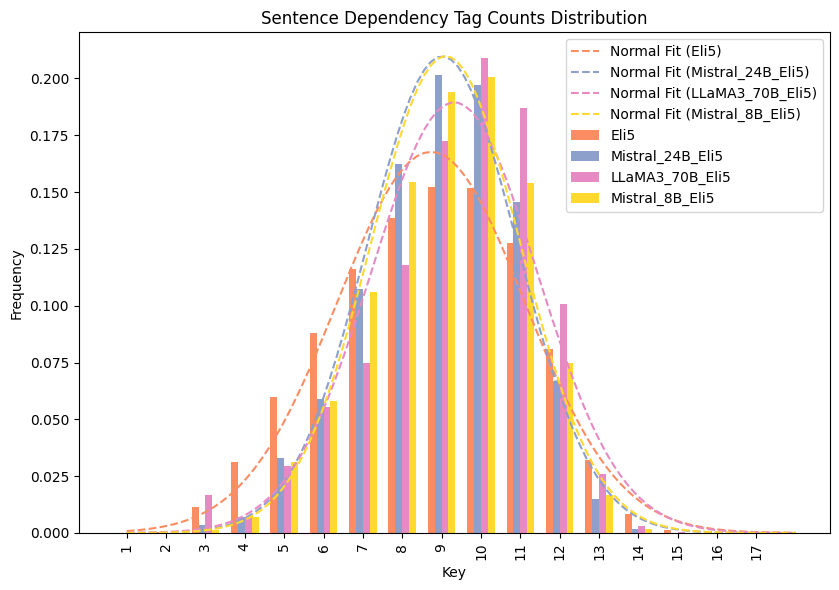}
 \caption{Proportion of sentences in ELI5 that have a particular number of unique dependency tags. }
\label{fig:deptageli5}
\end{figure}

As Figures~\ref{fig:deptagccnews}--\ref{fig:deptageli5} show, the regenerated data in each domain is more narrow and has a slightly upward shifted mean relative to the human data. The original human data is close to normally distributed, so we do not report a reduced long tail for this metric.




\subsection{Depth Scores}
\begin{figure}[ht]
    \centering
    \includegraphics[width=1\linewidth]{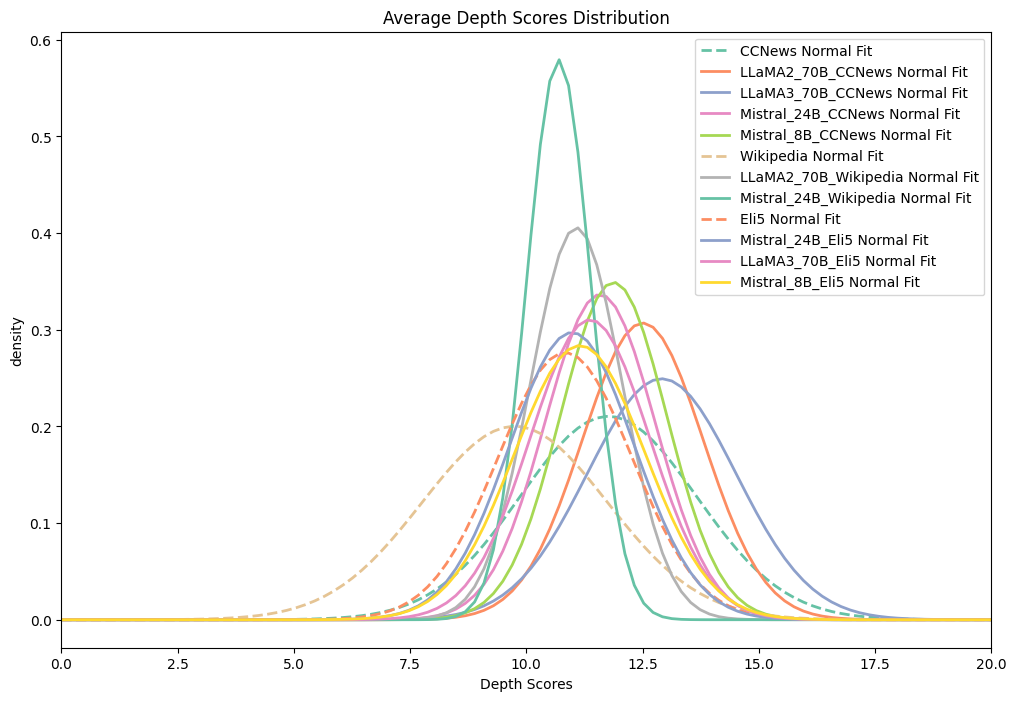}
    \caption{Average depth score normal fits. Dotted lines indicate human original domains, and solid lines indicate model regenerations.}
    \label{fig:average_depth_score_normal_fits}
\end{figure}

Normal fits for depth scores are provided in Figure~\ref{fig:average_depth_score_normal_fits}. For Wikipedia, and CCNews, the parse tree depth of the regenerated data shows a narrowing effect. Interestingly, the regenerated data for ELI5 is close in variance to the original data across models. This is the only time in our study that we do not find a clear narrowing effect. For the \llama-70B-regenerated data, we find a higher mean across all domains. Across domains and models (with the exception of Mistral-24B's CCNews and Wikipedia data), the slight right tail is reproduced in the regenerated data (see Figure~\ref{fig:depthscores} in Appendix~\ref{app:depth}).

\subsection{Yngve Scores}

\begin{figure}[!ht]
    \centering
    \includegraphics[width=1\linewidth]{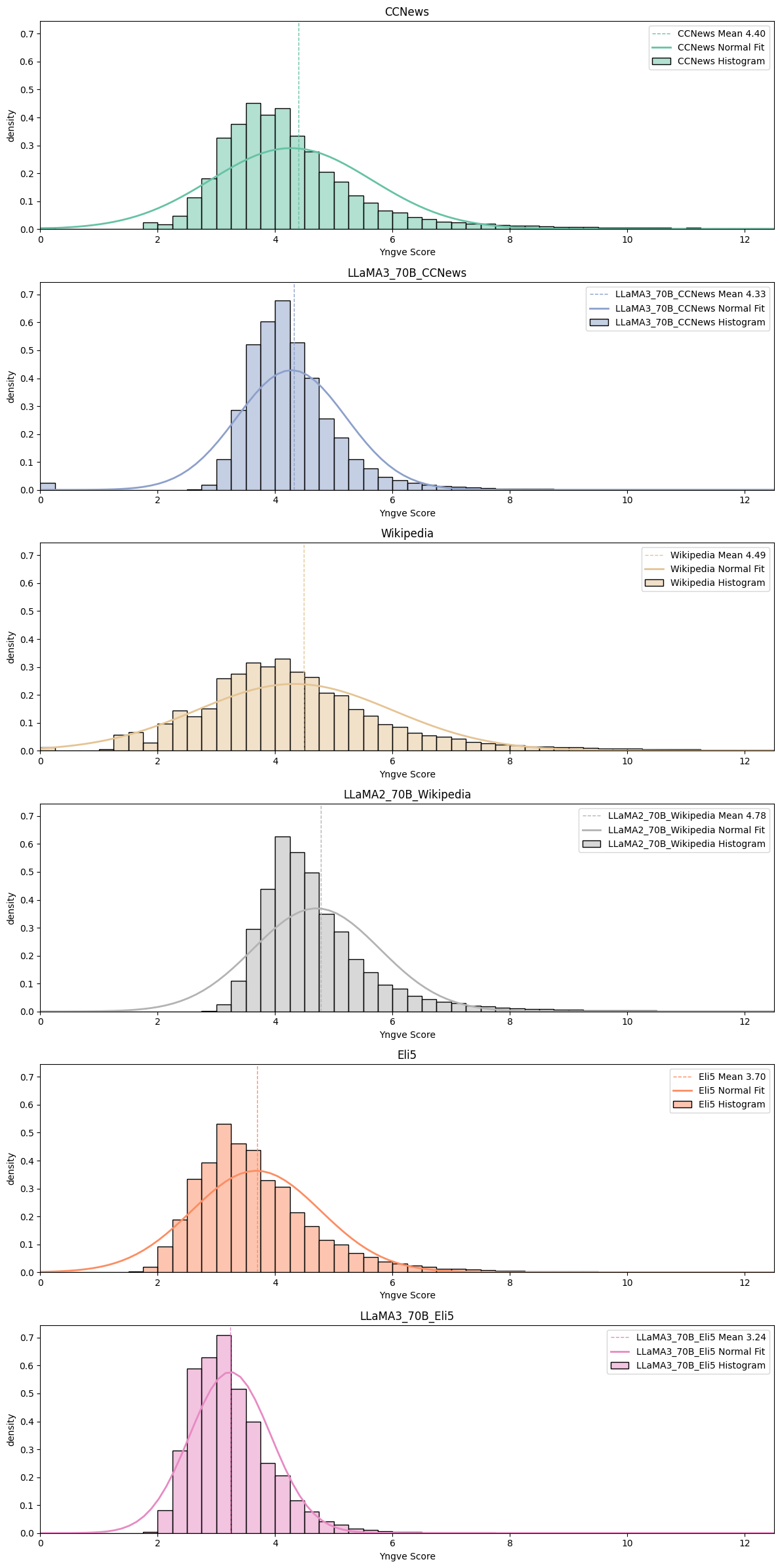}
    \caption{Yngve parse complexity score distribution for each dataset. Normal distribution curves fit to the data show that both human and \llama-regenerated datasets are not particular normally distributed. \llama-regenerated datasets show a narrower distribution than human with a heavy right tail that is reduced in comparison to the human datasets (but still visibly present).}
    \label{fig:yngvescores}
\end{figure}

Yngve scores for \llama-regenerated data are shown in Figure~\ref{fig:yngvescores}. In all three domains, we see considerably more narrow distributions in the \llama-generated data, and a shorter, less heavy long right tail. Nonetheless, the \llama-regenerated data shows a considerable right tail. The mean is increased relative to the human mean in the case of Wikipedia, while in the CCNews data, and ELI5, the mean is slightly lower. In Appendix~\ref{app:yngve}, we plot the full results for all tested models in \autoref{fig:Yngve-Score-full}, where data regenerated with other models show the same trends.

\subsection{Constituency Labels} 
\begin{figure*}[h]
\centering
\includegraphics[max width = \textwidth]{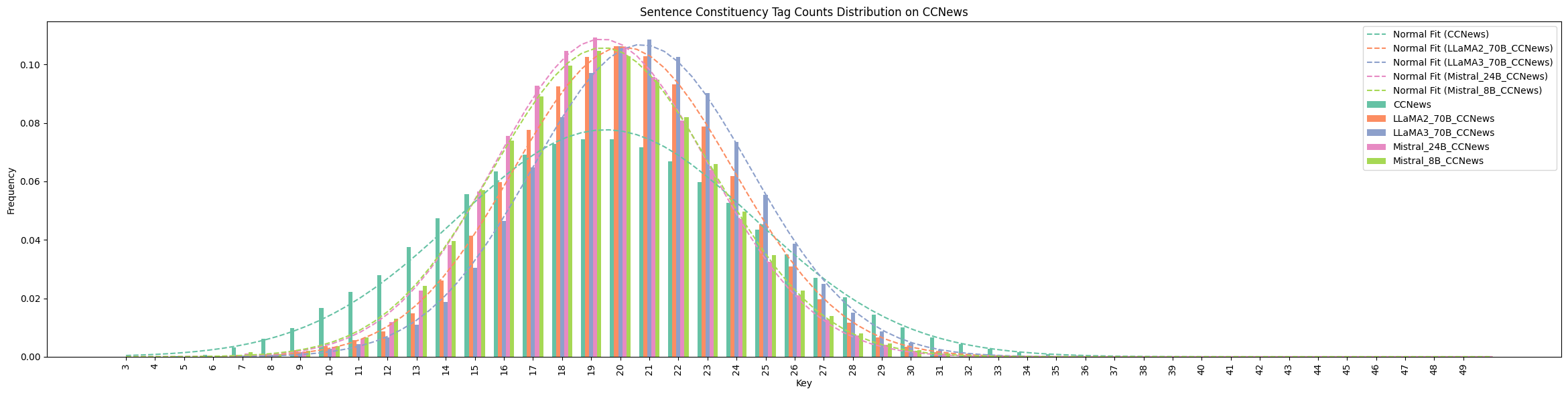}%
   \caption{Proportion of sentences in CCNews (y-axis) that have a particular number of unique constituency labels (x-axis). Colors indicate whether the distribution belongs to the original source data, or the source domain as regenerated by \llama-V2 or \llama-V3.
   }
   \label{fig:constituencyccnews}%
\end{figure*}

\begin{figure*}[h]
\includegraphics[max width = \textwidth]{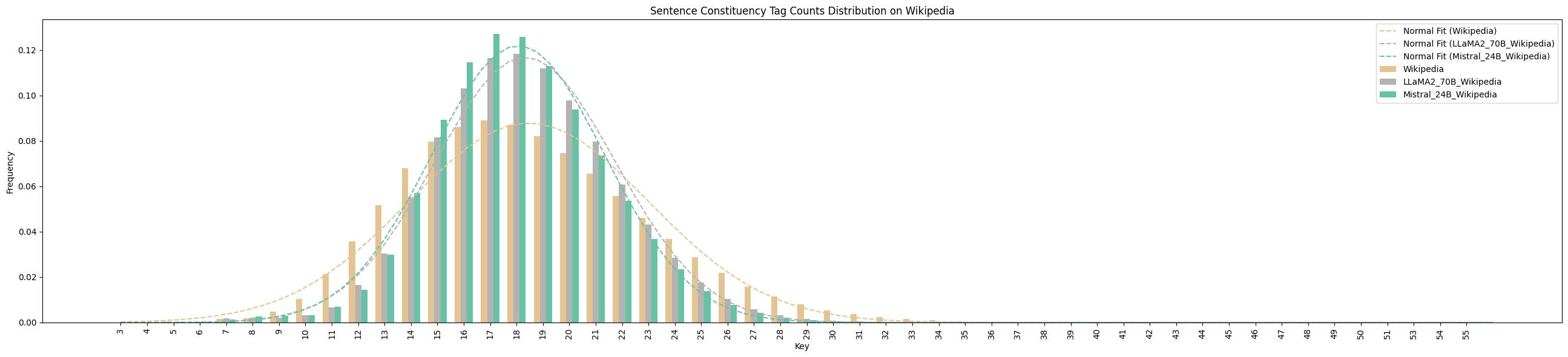}
 \caption{Proportion of sentences in Wikipedia (y-axis) that have a particular number of unique constituency labels (x-axis). Colors indicate whether the distribution belongs to the original source data, or the source domain as regenerated by \llama-V2 or \llama-V3.
 }
\label{fig:constituencywiki}
\end{figure*}

\begin{figure*}[!h]
\includegraphics[max width = \textwidth]{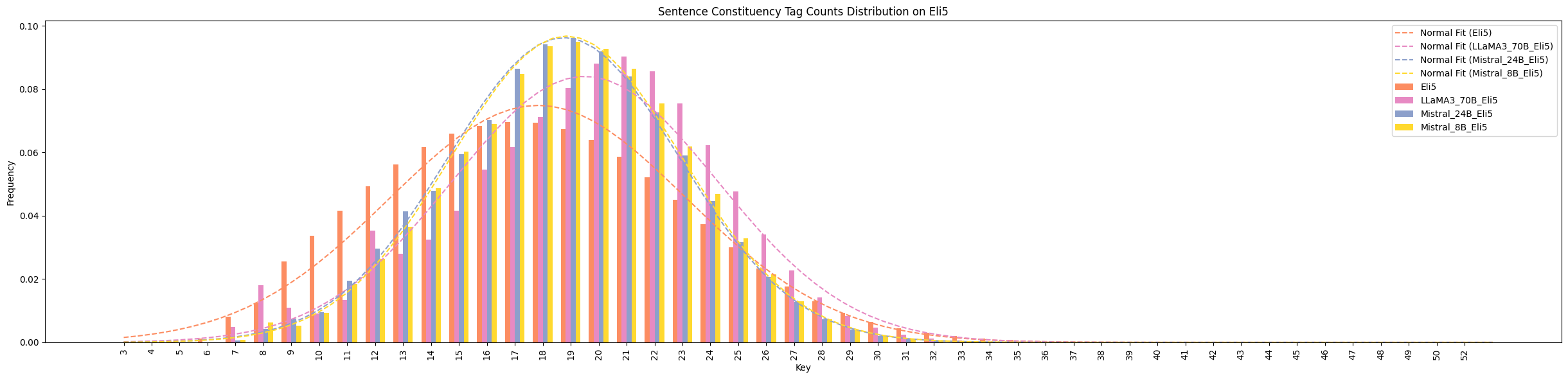}
 \caption{Proportion of sentences in ELI5 (y-axis) that have a particular number of unique constituency labels (x-axis). }
\label{fig:constituencyeli5}
\end{figure*}

Figures~\ref{fig:constituencyccnews}--\ref{fig:constituencyeli5} show the distribution of unique constituency labels per sentence. Across domains and models, we find that the regenerated data has a narrower distribution. The human distributions exhibit a slight right tail that is largely absent with in the regenerated data. In the regenerated CCNews, the smaller Mistral models have a slightly downshifted mean, while the \llama~models shift their mean upwards. For Wikipedia, and ELI5, the models all shift their mean upwards.


\subsection{Summary and Interpretation of Results}

Across the majority of our metrics and datasets, we observe that models generate distributionally less diverse data, as evidenced by both a decrease in variability and a reduced long tail. Both of these signatures are compatible with the interpretation that the models are simplifying relative to the human domain: if they were generating syntactically simpler sentences overall, this could be underlying the lower variability, and if they were failing to capture rarer syntactic phenomena, or rarer combinations of syntactic phenomena, this might account for the reduced long tail.

For example, the long tail of Yngve scores (\autoref{fig:yngvescores}) would contain strongly left branching structures that are grammatical but rare in English (e.g., clausal subjects). Many linguists take such structures to be evidence of the recursive nature of the combinatorial system that underlies human languages, and as such a loss of the long tail is compatible with syntactic simplification by the LLMs.

Interestingly, even for the unique dependency tags metric, where we report no long tail reduction, because the human data is very close to normally distributed, the regenerated data deviates from the human data. Across our other metrics, the human data is more left-leaning (exhibits a right tail) than the regenerated data. Even here, where the human data is very close to normal, the regenerated data leans right of that, thus showing the same directional difference as elsewhere.

However, while the model-regenerated data is uniformly less diverse than the corresponding human-generated texts across our variety of measures (reduced variability and a reduced long tail), the mean shifts vary across domains: For CCNews and Wikipedia, the \llama-generated data is mostly shifted upwards, while the mean of the regenerated ELI5 tends to match or be shifted downwards relative to the original data. Since ELI5 is presumably more simple than the overall average training data (while CCNews and Wikipedia may either match it or be more complex), the direction of mean shifts in the regenerated data suggests that models overshoot their domain complexity. In other words, in terms of the mean of our complexity measures (but not in terms of diversity) the models appear to regenerate domain data that is a more extreme form of the human original (simplifying in the case of the simpler ELI5 domain, and shifting the mean complexity upwards in the case of Wikipedia and CCNews).

What we don't see is all regenerated data consistently landing at some kind of middle ground---a ``neutral'' domain, that would reflect the properties of some average of all its training data---regardless of which domain the model is prompted to match. Since this isn't present, we can conclude that the models do have some notion of domain and encode the fact that domains differ. However, the data that models regenerate are consistently less diverse, and in terms of the mean of our complexity measures, they tend to be more extreme than the human originals, suggesting that the models' notion of domain is not fully humanlike.

\section{Qualitative Reflections on LLM-Regenerated Data}

\paragraph{\llama-V2 Regenerated Wikipedia.} We observed several trends when manually inspecting the data regenerated by the LLMs.~When comparing \llama-V2-regenerated Wikipedia articles to the original human ones, we observed spelling normalizations (e.g.\ British \textit{-ise} becomes American \textit{-ize}), increased inclusion of value judgments (which go against Wikimedia's editorial guidelines called the ``Neutral Point of View''\footnote{\href{https:\\meta.wikimedia.org/wiki/Neutral_point_of_view}{https://meta.wikimedia.org/wiki/Neutral\_point\_of\_view}})---and an increased prevalence of essay-like wrap-up sentences. 

For example, the final sentences of the \llama-V2 regenerated article on ``A'' 
are both explicitly concluding and unusually complimentary:
\textit{In conclusion, the letter \emph{a} is an important and versatile letter in the English language. It is used as an indefinite article, a pronoun, a prefix, a suffix, and in many abbreviations and acronyms.}\footnote{Note also that it wrongly asserts that the letter can be used as a pronoun (a property that indefinite articles like English ``a'' do indeed have in other languages like German, but not in English) and a suffix, of which the text alleges elsewhere that it denotes the performer of an action (peculiarly, in non-rhotic variants of English, the Latinate plural \textit{-a} may be homophonous with the agent nominalization \textit{-er}, which does indeed denote the performer of an action).} As may be clear from this example, the models introduce stylistic elements that are not generally in keeping with the style of the original Wikipedia domain. More examples are provided in Appendix~\ref{appendix:wikiexamples}. These observations point the way to future work that explores the consequences of the domain regeneration paradigm on stylistic elements.

\paragraph{\llama-Regenerated CCNews.}

\llama-V2 and \llama-V3 were both prone to inserting a higher number of quotations attributed to famous or influential people than the original articles, which largely described an event. For example, we saw novel inclusions of quotations attributed to Jeff Gundlach, Warren Buffet, Mark Schneider, Zhang Yuhua, Chen Qi, and David Cameron. In the rare case where the original CCNews articles contained quotations, usually only one quote was present. It would be fairly long in comparison to \llama-V2 regenerated CCNews quotes. There were also a number of cases of \llama-V2 inserting acronyms where none had existed in the source (e.g. \textit{Albuquerque Little Theatre (ALT)}).

CCNews regenerated by \llama-V2 and \llama-V3 also displayed unusual wrap-up sentences, except, unlike for Wikipedia, they appeared to be more PR or sales related. For example, the article on ``ARKit 1.5'' ended with \textit{Whether you're a tech enthusiast, a developer, or simply someone interested in the future of technology, ARKit 1.5 demos are certainly worth keeping an eye on.} More examples and comparisons between \llama-V2 and \llama-V3 are present in the Appendix~\ref{appendix:ccnewsexamples}.

\section{Related Work}

Closest to our work is \citet{shaib-etal-2024-detection}, which explored sequences of part-of-speech tags in training data and model generations. They reported that several LLMs generated more syntactically homogeneous text, as compared to human ground truth. Our work differs from theirs in that we focus on different datasets and models, and perform distinct experiments. We take a distributional view and are interested in the domain match setting, exploring additional signatures of model-and-human difference and more syntactic metrics. They explore neutral domain text, diving deeper into the effect of decoding temperature, and also exploring the additional summarization setting.  

\section{Conclusion}
Using our regeneration paradigm, we have uncovered systematic syntactic differences between human-generated and model-regenerated text. Across a variety of syntactic complexity metrics, the regenerated text showed lower variability as well as a reduced long tail, when compared against the human-generated text in the same domain, while the mean of the measurements was often shifted in a way that suggests that models overshoot when trying to match domain properties.

Our results may have practical implications (i.e. on decisions about whether or not to utilize LMs as components in domain transfer systems) and theoretical implications (e.g. about the empirical status of syntactic long tail effects).

\section*{Acknowledgments}
We would like to thank Chantal Shaib, Jessica Forde, Candace Ross, and Levent Sagun for conversations relating to model collapse. We'd also like to thank Sebastian Ruder for feedback on an early draft, and the audience of the Organized Session on LLMs, Linguistics, and Psycholinguistics at the 2025 Annual Meeting of the Linguistic Society of America in Philadelphia, PA for comments and questions that helped us shape the final paper.

\section{Limitations}

\paragraph{Tooling and Pipeline.} While our visual inspections didn't surface any immediate issues, we acknowledge the possibility of tooling failures when we try to calculate metrics or parse sentences that are extremely long or complicated. However, since we uniformly apply our tools across domains and generation sources, we expect any errors to be comparable, and thus not to have an outsized impact on our results. 

\paragraph{Decoding Temperature.} In this work, we used the default temperature from vLLM. We presume that lowering the temperature would decrease randomness, presumably further reduce diversity, and  higher temperature could increase diversity, but it is not immediately clear what the effect would be on the reduction of the long tail. A more thorough exploration of decoding temperature could be explored in future work. 

\paragraph{Syntactic Complexity Metrics.} In this work, we utilized existing complexity metrics from prior literature. However, we have anecdotally observed some additional changes to the style and content, which one could devise metrics to specifically target. Future work could perform more data analysis to help guide the creation of additional informative syntactic complexity metrics, which, in turn, could help us gain more insights into the type of simplification LLMs affect, and inspire architectural or training improvements.  

\bibliography{cited}

\clearpage
\appendix

\section{Ablation of data cleaning}
\label{appendix:ablation_data}
In this section, we also present our results for less filtered data on two of our domains, Wikipedia and CCNews. Overall, we see the same trends as for the filtered data presented in the main paper. For dependency tags and constituency labels for both datasets: we see mean shift (mostly for CCNews) and narrowing (for both datasets). Descriptive statistics following the length and POS filtration are presented in Table \ref{tab:descriptive_statistics}. Our overall cleaning and processing pipeline (including length filtering) excluded on average less than 10\% in the case of CCNews datasets, and less than 15\% in the case of Wikipedia datasets.
\begin{table}[ht!]
\begin{minipage}{0.5\textwidth}
\resizebox{0.9\columnwidth}{!}{%
\begin{tabular}{lccccc}
\toprule
Datasets  & Sentences & Words & S/A & W/S & W/A \\
\midrule
CCNews    & 13.8M     & 0.3B  & 23.9 & 23.2 & 554.3 \\
LLaMA2 70B    & 20.6M     & 0.5B  & 29.1 & 24.9 & 724.9 \\
LLaMA3 70B    & 24.5M     & 0.7B  & 34.6 & 27.3 & 946.4 \\
\midrule
Wikipedia & 129.5M    & 2.9B  & 20.0 & 22.5 & 450.4 \\
LLaMA 2 70B   & 277.5M    & 5.6B  & 42.0 & 20.3 & 854.9 \\
\bottomrule
\end{tabular}}
\caption{Descriptive statistics on raw data.}
\end{minipage}
\begin{minipage}{0.5\textwidth}
\resizebox{0.9\columnwidth}{!}{%
\begin{tabular}{lrrrrrr}
\toprule
Datasets  & Sentences & Words & S/A & W/S & W/A \\
\midrule
CCNews  & 13.6M & 0.3B & 23.4 & 23.5 & 551.3 \\
LLaMA2 70B & 20.4M & 0.5B & 28.8 & 25.1 & 724.4 \\
LLaMA3 70B & 24.4M & 0.7B & 34.5 & 27.4 & 946.2 \\
\midrule
Wikipedia  & 122.8M & 2.9B & 19.0 & 23.6 & 448.1 \\
LLaMA 2 70B & 257.3M & 5.6B & 39.0 & 21.8 & 850.2 \\
\bottomrule
\end{tabular}}
\caption{Descriptive statistics after length filtration.}
\end{minipage}
\end{table}

Readability scores were consistently calculated on entire articles without any data cleaning. We compute readability scores for articles containing more than 100 words, as shorter articles do not provide sufficient indicators of readability. 
Metrics such as the depth score and Yngve score, which were initially aggregated at the article level, showed minimal variation upon inspection. Therefore, they are not included in the ablation results presented here.
For reference, we provide results on dependency and constituency parsing using the raw data below.
\begin{figure*}[!ht]
\centering
\begin{minipage}[t]{0.47\textwidth}
\includegraphics[max width = \textwidth]{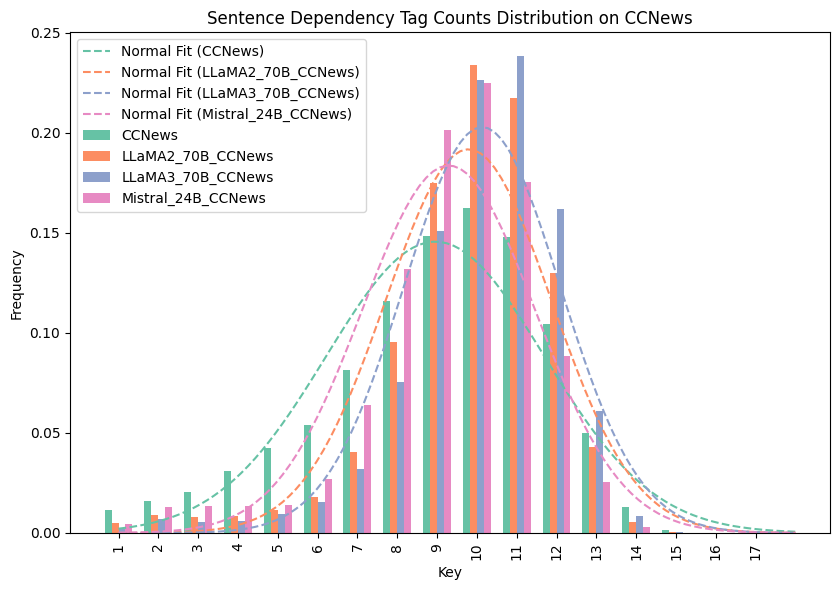}%
   \caption{Proportion of sentences in CCNews (y-axis) that have a particular number of unique dependency tags (x-axis). Colors indicate whether the distribution belongs to the original source data, or the source domain as regenerated by \llama-V2 or \llama-V3. This comparison was made on data without any cleaning.} 
   \label{fig:raw_deptagccnews}%
\end{minipage}%
\hfill
\begin{minipage}[t]{.47\textwidth}
\includegraphics[max width = \textwidth]{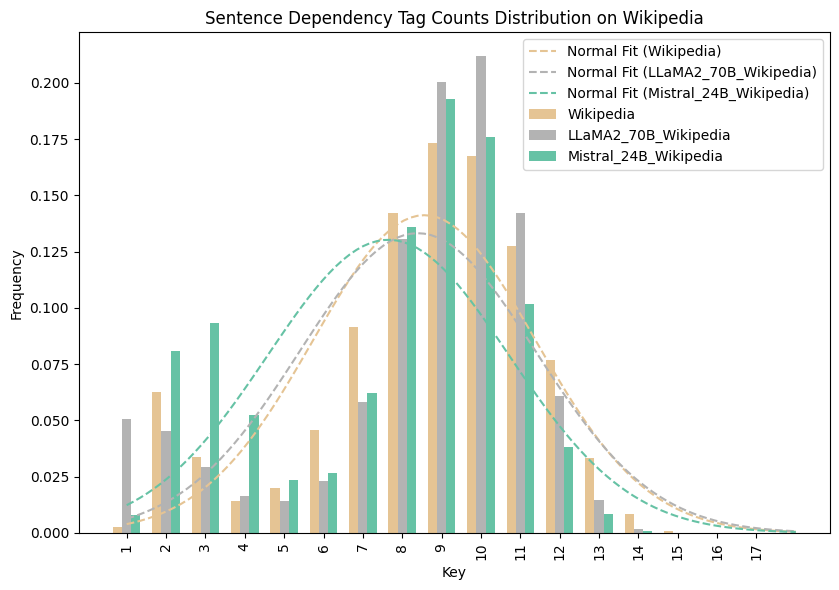}
 \caption{Proportion of sentences in Wikipedia (y-axis) that have a particular number of unique dependency tags (x-axis). Colors indicate whether the distribution belongs to the original source data, or the source domain as regenerated by \llama-V2 or \llama-V3. This comparison was made on data without any cleaning.}
\label{fig:raw_deptagwiki}
\end{minipage}
\end{figure*}
\begin{figure*}[t]
\centering
\begin{minipage}[t]{\textwidth}
    \centering
    \includegraphics[max width=\textwidth]{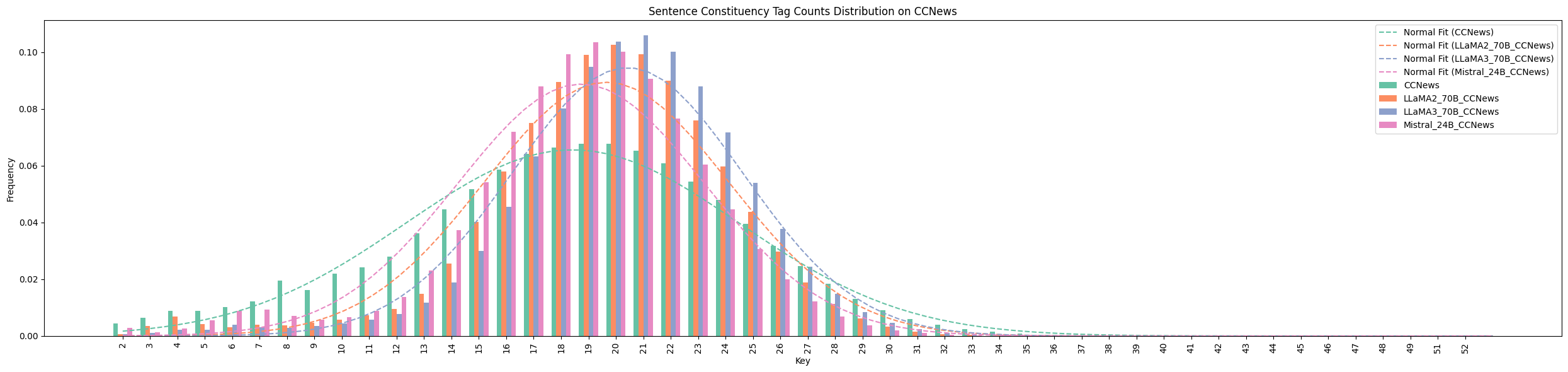}
    \caption{Proportion of sentences in CCNews (y-axis) that have a particular number of unique constituency tags (x-axis). Colors indicate whether the distribution belongs to the original source data, or the source domain as regenerated by \llama-V2 or \llama-V3. This comparison was made on data without any cleaning.}
    \label{fig:raw_constituency_ccnews}
\end{minipage}

\vspace{0.5cm} 

\begin{minipage}[t]{\textwidth}
    \centering
    \includegraphics[max width=\textwidth]{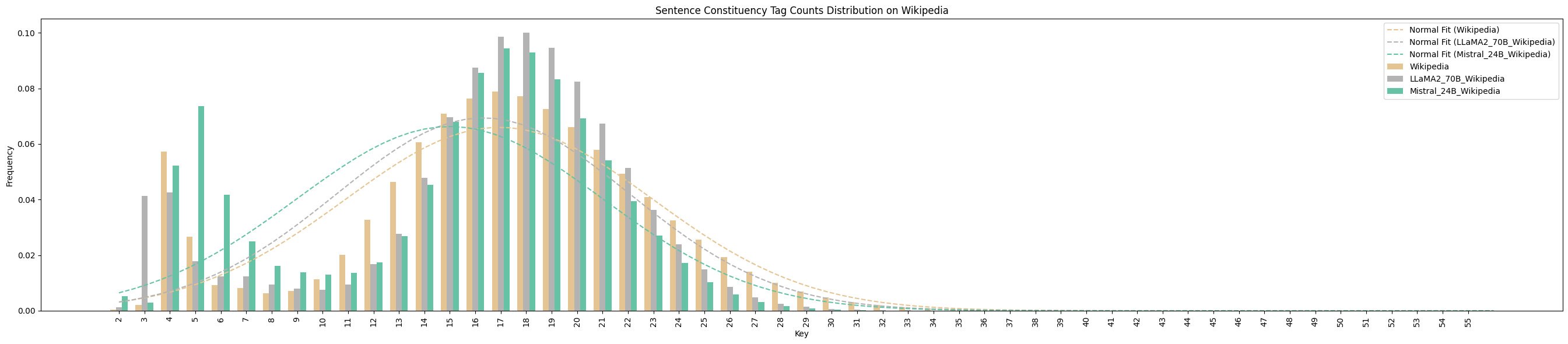}
    \caption{Proportion of sentences in Wikipedia (y-axis) that have a particular number of unique constituency tags (x-axis). Colors indicate whether the distribution belongs to the original source data, or the source domain as regenerated by \llama-V2 or \llama-V3. This comparison was made on data without any cleaning.}
    \label{fig:raw_constituency_wiki}
\end{minipage}
\end{figure*}

\paragraph{Main Results Figures Ablations.} Figures~\ref{fig:raw_deptagccnews}--\ref{fig:raw_constituency_wiki} present data ablations on our main results. Overall, we see the same rough trends as described in the main paper. 

For \autoref{fig:raw_deptagwiki}, we can see a difference in the dependency parses for fewer than three words between the human-generated data and the \llama-V2-generated data. We decided to filter out these lengths, because we expected the data to be noisy and uninformative about syntactic structure (there are very parses available for complete sentences with 3 or fewer words). We observe a similar trend for \autoref{fig:raw_constituency_wiki}, except that the noisy section extend to approximately 6 constituency labels, which is reflective of the same underlying fact that there are not many available parses for 3 words.

\section{Prompts}\label{appendix:prompts}
We prompted the LLMs using the two prompts below, one for each text domain. We retrieved the \{topic\} and \{title\} respectively from the original data sources and fed them into the model prior to including the initial section of text for the model to attempt to regenerate. Next, we included instructions that match standard instruction-tuning prompts, which also included a target article length in words, which we set to the average length of article from each domain. We observed that the regenerations were never word-for-word identical to the human versions, and we also observed that the models did not generate exactly the average lengths provided in the prompts.

\subsection{Wikipedia}\label{apptab:wikiprompt}
\lstset{
    basicstyle=\ttfamily\small,
    breaklines=true,
    frame=single,
    numbers=left,
    numberstyle=\tiny,
    stepnumber=1,
    numbersep=5pt,
    showstringspaces=false,
    tabsize=2,
    captionpos=b,
    breakatwhitespace=false,
    keywordstyle=\color{blue},
    commentstyle=\color{gray},
    stringstyle=\color{red},
    xleftmargin=0pt,
    xrightmargin=0pt,
    linewidth=\columnwidth
}
\begin{lstlisting}
NUM_FIRST_PARA_LENGTH = 256
TEXT_PROMPT = """
Generate a Wikipedia article on the topic of {topic}. 
Use the following first paragraph from the original Wikipedia article as a starting point:

{first_paragraph}

Now, expand upon the provided paragraph by providing additional details, 
historical context, notable events, key figures, and any relevant subtopics. 
Aim for a well-structured and informative Wikipedia style article with a minimum length of 700 words. 
Ensure that the content is factually accurate, well-written, and on Wikipedia writing style.
"""
\end{lstlisting}
\subsection{CCNews}\label{apptab:ccnewprompt}
\begin{lstlisting}
NUM_FIRST_PARA_LENGTH = 180
TEXT_PROMPT = """
Generate a news article on the topic of {title}.
Use the following first paragraph from the original news article as a starting point:

{first_paragraph}

Now, expand upon the provided paragraph by providing additional details, context, notable events, key figures, and any relevant discussions. Aim for a well-structured and informative news style article with a minimum length of 500 words. Ensure that the content is factually accurate, well-written, and on news writing style.
"""
\end{lstlisting}

\subsection{Eli5}\label{apptab:eli5_prompts}
\begin{lstlisting}
TEXT_PROMPT = """
Generate a reddit reply to this thread {title}.

Aim for an Explain Like I'm Five style reply with a minimum length of 100 words. Ensure that the content is factually accurate, well-written, and on Explain like I'm Five writing style.
"""
\end{lstlisting}

\section{Sentence Lengths}\label{app:length}

\begin{figure}[!h]
    \centering
    \includegraphics[width=1.5\linewidth,height=0.95\textheight,keepaspectratio]{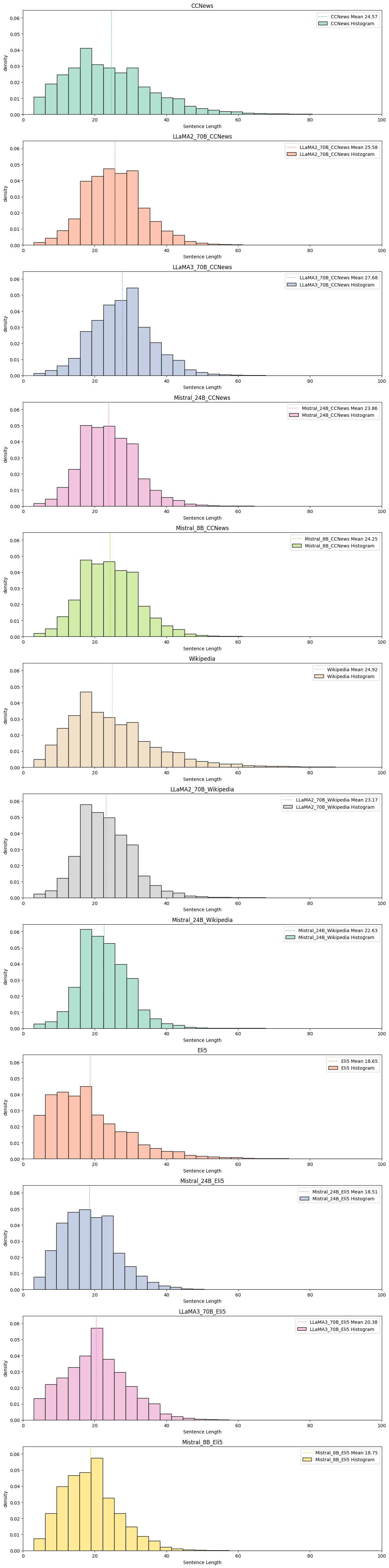}
    \caption{The full plot of sentence length for all the models.}
    \label{fig:Sentence-Length-full}
\end{figure}

As Figure~\ref{fig:Sentence-Length-full} illustrates, no length distribution is perfectly normal. When compared to the \llama-70B regenerations, We observe that the original human distributions have shorter sentences on average (i.e., the regenerated distributions have upward shifted means) for for CCNews and ELI5, but longer sentences for Wikipedia. We also observe that the original human distributions are also wider (i.e., the regenerated distributions have less variance) for all data sources (with ELI5 being the weakest effect, likely because more generations are at the length floor). Finally, we observe that the original distributions appear to have a longer and heavier right tail than their model generated counterparts for all data sources. For the other models, the mean shift is inconsistent across models, but all models show reduced variability, and a reduced long tail.

\section{More Information on Readability}
\label{appendix:additional_readability}

\subsection{Full results for the Flesh-Kincaid Readability Scores for all tested models.}

\begin{figure}[!ht]
    \centering
\includegraphics[width=1.5\linewidth,height=0.95\textheight,keepaspectratio]{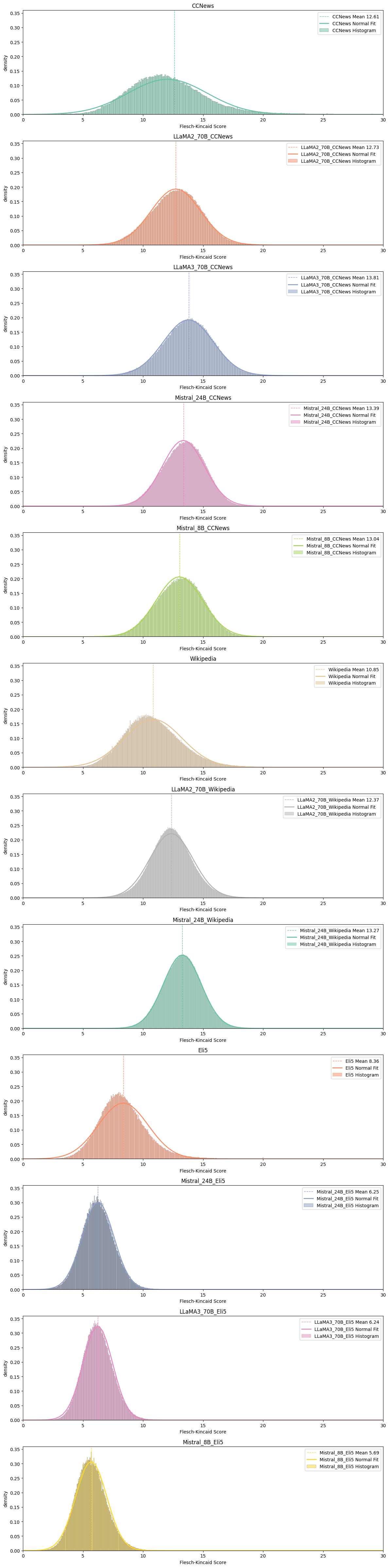}
    \caption{The full plot of Flesch-Kincaid Readability for all the models.}
    \label{fig:Flesch-Kincaid-full}
\end{figure}

We present full results for Flesh-Kincaid scores in \autoref{fig:Flesch-Kincaid-full}.

\paragraph{CCNews.} For the CCNews datasource, all model regenerated datasets have upward shifted means and narrowed distributions, in keeping with the subset presented in the main paper. They are all additionally more left than right tailed, when compared to the original human distribution.

\paragraph{Wikipedia.} For the Wikipedia datasource, \llama-V2-70B and Mistral-24B both had upward shifted mean, a narrower distribution, and a reduced right tail. Interestingly, the Mistral model has a near-perfectly normal distribution, while \llama-V2-70B retains a slight right tail. 

\paragraph{ELI5.} For the ELI5 datasource, all model-regenerated distributions have a downward shifted mean, a narrower distribution and a strongly reduced right tail. 

\subsection{Additional Readability Scores}

We also report the means, medians, and standard deviations of several readability scores, including Flesch-Kincaid grade level (see \autoref{fig:fkgrademeans}), for Wikipedia and CCNews domain data. We are using all 70B models for this analysis. We expect most of these metrics to correlate highly---as all of them are based on different combinations of words per sentence and syllables per word---but are including them in case they may be of interest to some readers.  Across all readability metrics (means and medians), human-generated Wikipedia is deemed the simplest datasets, and \llama-V3-CCNews is deemed the most complex. Standard deviations are generally within the same range for all metrics, except for human-generated CCNews which has the most variation in readability. Second in highest standard deviations across scores is human-generated Wikipedia (Flesch-Kincaid Reading Ease, Linsear Write) or \llama-V3-generated CCNews (Gunning-Fog Index, Spache Readability Formula, Linsear Write). The fact that human-generated data has high standard deviations across the board (although occasionally in a tie with \llama-V3-CCNews) suggests the human distributions have more diversity in generations. 

\paragraph{Flesch-Kincaid Reading Ease.} We report the Flesch-Kincaid Reading Ease metric, which is similar to the Flesch-Kincaid grade level scoring in that it is calculated from number of words and syllables per sentence. A higher score indicates that the material is easier to read. Human-generated Wikipedia has the highest reading ease ($50$), and \llama-V3-generated CCNews has the lowest ($39.5$), but both fall into the range expected for college level texts. The Flesch-Kincaid Reading Ease are presented in \autoref{fig:fkreadingeasemeans}.

\paragraph{Gunning-Fog Index.}
The Gunning-Fog index \citep{gunning-1952-technique} is another estimate of reading level, which is also based on the number of words per sentence and the number of syllables per word, but it generally has a lower value than the Flesh-Kincaid grade level and reading ease scores. The Gunning-Fog scores are presented in \autoref{fig:gfmeans}. 

\paragraph{Linsear Write Scores.} The Linsear Write Scores are something of a thresholded version of the other scores, where the words with more syllables are deemed ``challenging'' and words with fewer syllables are deemed ``easy''. The Linsear Write Scores are presented in \autoref{fig:linsearmeans}.

\paragraph{Spache Readability Formula.}
The Spache Readability Formula \citep{spache-1953-new} operates on a list of words that are expected to be familiar for children up until the fourth grade in the United States. The formula considers average sentence length and proportion of familiar words to determine its score. Of all the metrics reported, this score resolves the least differences between datasets. The Spache Readability Formula scores are presented in \autoref{fig:spachemeans}.

\begin{table}[!ht]
\resizebox{\columnwidth}{!}{%
\begin{tabular}{lrrrr}
\toprule
Dataset  & Mean & Median & STD & Sample Size \\
\midrule
CCNews & 12.6 & 11.8 & 8.2 & 561167 \\
\llama-V2 & 12.7 & 12.8 & 2.1 & 708011 \\
\llama-V3 & 13.8 & 13.8 & 3.2 & 702530 \\
Wikipedia & 10.8 & 10.7 & 2.7 & 3829535 \\
\llama-V2 & 12.4 & 12.3 & 2.1 & 6601865 \\
\bottomrule
\end{tabular}}
\caption{Flesch-Kincaid Grade Level}\label{fig:fkgrademeans}
\end{table}

\begin{table}[!ht]
\resizebox{\columnwidth}{!}{%
\begin{tabular}{lrrrr}
\toprule
Dataset  & Mean & Median & STD & Sample Size \\
\midrule
CCNews & 48.2 & 50.4 & 24.9 & 561167 \\
\llama-V2 & 43.4 & 43.2 & 11.3 & 708011 \\
\llama-V3 & 39.5 & 39.4 & 12.8 & 702530 \\
Wikipedia & 50.0 & 51.1 & 12.8 & 3829535 \\
\llama-V2 & 40.8 & 41.0 & 10.7 & 6601865 \\
\bottomrule
\end{tabular}}
\caption{Flesch-Kincaid Reading Ease}\label{fig:fkreadingeasemeans}
\end{table}

\begin{table}[!ht]
\resizebox{\columnwidth}{!}{%
\begin{tabular}{lrrrr}
\toprule
Dataset  & Mean & Median & STD & Sample Size \\
\midrule
CCNews & 14.4 & 13.5 & 8.5 & 561167 \\
\llama-V2 & 15.2 & 15.3 & 2.3 & 708011 \\
\llama-V3 & 16.5 & 16.5 & 3.5 & 702530 \\
Wikipedia & 12.0 & 11.7 & 3.0 & 3829535 \\
\llama-V2 & 14.2 & 14.2 & 2.4 & 6601865 \\
\bottomrule
\end{tabular}}
\caption{Gunning-Fog Index}\label{fig:gfmeans}
\end{table}

\begin{table}[!ht]
\resizebox{\columnwidth}{!}{%
\begin{tabular}{lrrrr}
\toprule
Dataset  & Mean & Median & STD & Sample Size \\
\midrule
CCNews & 16.6 & 15.1 & 14.1 & 561167 \\
\llama-V2 & 15.9 & 15.9 & 2.6 & 708011 \\
\llama-V3 & 17.6 & 17.5 & 4.3 & 702530 \\
Wikipedia & 12.9 & 12.7 & 4.2 & 3829535 \\
\llama-V2 & 14.3 & 14.3 & 3.3 & 6601865 \\
\bottomrule
\end{tabular}}
\caption{Linsear Write Scores}\label{fig:linsearmeans}
\end{table}

\begin{table}[!ht]
\resizebox{\columnwidth}{!}{%
\begin{tabular}{lrrrr}
\toprule
Dataset  & Mean & Median & STD & Sample Size \\
\midrule
CCNews & 8.1 & 7.8 & 3.2 & 561167 \\
\llama-V2 & 7.7 & 7.7 & 0.7 & 708011 \\
\llama-V3 & 8.0 & 8.0 & 1.2 & 702530 \\
Wikipedia & 7.6 & 7.6 & 1.0 & 3829535 \\
\llama-V2 & 7.6 & 7.6 & 0.7 & 6601865 \\
\bottomrule
\end{tabular}}
\caption{Spache Readability Formula Scores}\label{fig:spachemeans}
\end{table}




\section{Constituency Parse Depths Scores}\label{app:depth}
\begin{figure}[h]
    \centering
    \includegraphics[width=1.5\linewidth,height=0.95\textheight,keepaspectratio]{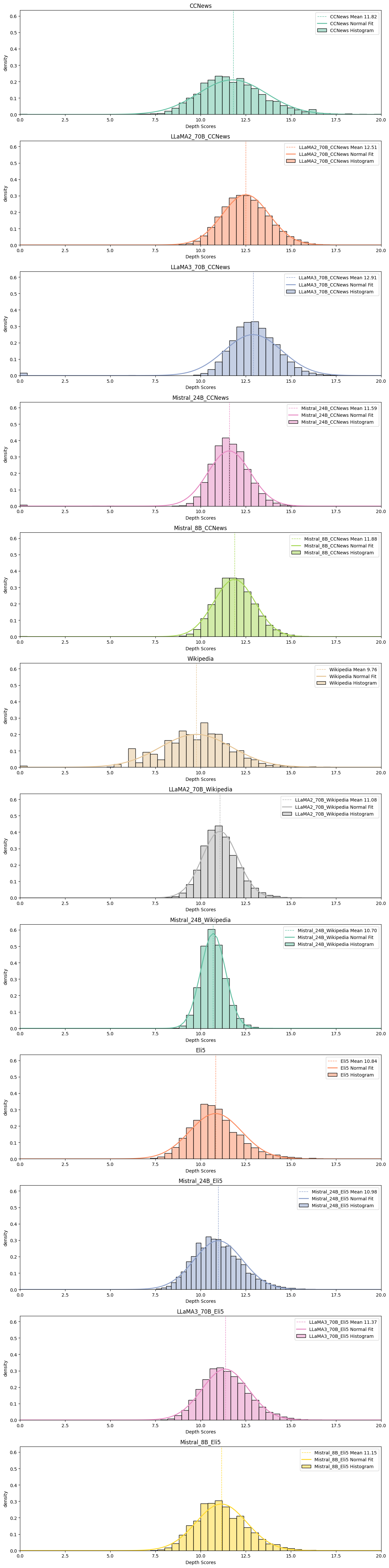}
    \caption{The distribution of constituency parse depth for each domain and datasource (human or model). Normal distribution curves show that \llama-regenerated datasets have a narrower distribution.}
    \label{fig:depthscores}
\end{figure}

\autoref{fig:depthscores} shows the distribution of constituency parse depths. Depth scores for model-regenerated text have an upwardly shifted mean, and a more narrow distribution when compared to human-generated text.

\section{Full Yngve Score Distribution Plot}\label{app:yngve}
\begin{figure}[!h]
    \centering
    \includegraphics[width=1.5\linewidth,height=0.95\textheight,keepaspectratio]{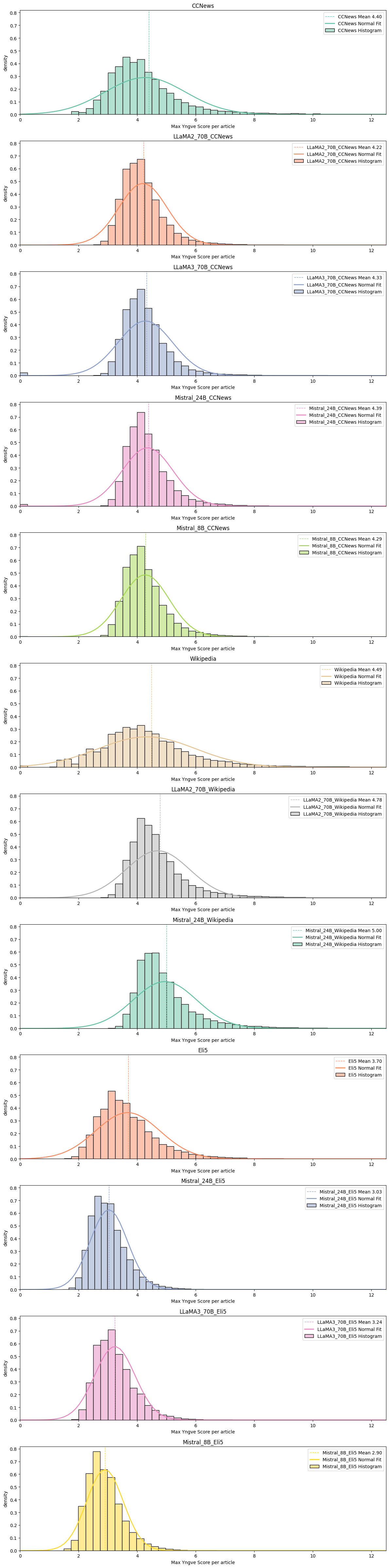}
    \caption{The full plot of Yngve score for all the models.}
    \label{fig:Yngve-Score-full}
\end{figure}

The full results for our Yngve metric on all models and domains can be found in \autoref{fig:Yngve-Score-full}.

\section{Example regenerations from \llama-V2 on the Wikipedia Domain}
\label{appendix:wikiexamples}

We manually inspected several examples from \llama-V2 and \llama-V3 and noticed some trends. Several of these trends are also listed in the \href{https://github.com/jenna-russell/human_detectors/blob/main/prompts/detection_guide.txt#L199}{supplementary materials} of \citet{russell2025peoplefrequentlyusechatgpt}, which gives tips and tricks for detecting model generated content; these include stylized and formalized conclusions, and spelling normalizations.  

\paragraph{Spelling Normalization.}
When the original article contained British spellings (e.g. \textit{-ise} as opposed to \textit{-ize}), \llama-2 did not uniformly preserve them, preferring American spellings for ``organize'', ``emphasize'', ``decentralize''. 

\paragraph{Longer snipped of the \llama-regenerated Wikipedia article ``A''.}
\textit{In addition to their use as indefinite articles, ``a'' and ``an'' can also be used as pronouns to refer to a noun or noun phrase that has been previously mentioned. For example: ``I saw a cat in the park.'' (meaning a specific cat that was seen in the park) ``I gave an apple to my friend.'' (meaning a specific apple that was given to a friend) In English, the letter ``a'' is also used as a prefix to indicate negation or absence, as in ``a-'' or ``non-''. For example: ``Atheist'' means someone who does not believe in God. ``Amoral'' means someone who does not have moral principles. ``Asexual'' means someone who does not experience sexual attraction. The letter ``a'' is also used as a suffix to indicate a noun that refers to a person or thing that performs a particular action or has a particular characteristic. For example: ``Artist'' means someone who creates art. ``Athlete'' means someone who is skilled in sports or physical activity. ``Actor'' means someone who performs in movies, television shows, or theater productions. The letter ``a'' is also used in many abbreviations and acronyms, such as ``A.M.'' (meaning ``ante meridiem'' or ``morning''), ``P.M.'' (meaning ``post meridiem'' or ``afternoon''), ``A.S.'' (meaning ``associate degree''), ``B.A.'' (meaning ``Bachelor of Arts''), ``M.A.'' (meaning ``Master of Arts''), and ``Ph.D.'' (meaning ``Doctor of Philosophy''). In addition to its use in English, the letter ``a'' is also used in many other languages, including Spanish, French, German, Italian, Portuguese, and many others. In these languages, ``a'' is used to represent a variety of vowel sounds, including the sounds /\textipa{a}/, /\textscripta/, /\textipa{@}/, and /\"{a}/. In conclusion, the letter ``a'' is an important and versatile letter in the English language. It is used as an indefinite article, a pronoun, a prefix, a suffix, and in many abbreviations and acronyms. Its origins can be traced back to the Phoenician alphabet, where it represented a glottal stop, and it has evolved over time to represent a variety of vowel sounds in different languages.}

\paragraph{Conclusion summaries and value judgments.} The regenerated article on ``Albedo'' concluded with \textit{By understanding albedo, scientists and policymakers can make informed decisions about environmental management, urban planning, agriculture, and other fields that affect human life and the environment.}

 ``Abraham Lincoln'': \textit{In conclusion, Abraham Lincoln was a strong leader who fought to preserve the Union and end slavery. His eloquence, political skills, and commitment to democracy and freedom have made him a beloved figure in American history. His legacy endures, and he remains an icon of American values and ideals.}

``Alabama'':
\textit{From its Native American roots to its current status as a thriving Southern state, Alabama has contributed to the nation's cultural, economic, and political landscape. Its people, events, and landmarks continue to inspire and captivate the nation's attention, and its legacy will undoubtedly endure for generations to come.}

``Achilles'':
\textit{Achilles' legacy continues to live on, and his story will continue to inspire and captivate audiences for generations to come.}

``Aristotle'':
\textit{Aristotle's legacy is a testament to his profound impact on human thought and culture, and his ideas will continue to shape our understanding of the world for generations to come.}

``Academy Awards'': \textit{Whether it's the red carpet glamour, the thrill of winning an Oscar, or the memories of unforgettable moments, the Academy Awards continue to capture the hearts and imaginations of people around the world.}

%

\section{Example regenerations from \llama-V2 on the CCNews Domain}\label{appendix:ccnewsexamples}

\paragraph{Conclusion summaries and sales pitches.}

The \llama-V2 regenerated article on ``China launches 1st X-ray Space Telescope'' concludes with \textit{In conclusion, the launch of China's first X-ray space telescope, `Insight', marks a significant milestone in the country's space exploration history. The telescope is expected to provide valuable insights into some of the most complex and fascinating phenomena in the universe and push the boundaries of detection technology. The launch is a testament to China's commitment to advancing the field of astronomy and its growing influence in the global space community.}

An article on Nestl\'{e} being optimistic that ingredient prices will fall concludes with \textit{In conclusion, Nestle's optimism about its ability to offset the impact of rising ingredient prices and meet its sales growth target is a positive sign for the company's future prospects. The company's diversified portfolio, geographic presence, and commitment to sustainability will help to mitigate any negative impact from economic uncertainty and drive long-term growth. As the company continues to invest in premium and healthier products, as well as sustainable sourcing, it is well-positioned to meet the changing needs of consumers and deliver on its growth targets.}

``Magical Show Mary Poppins to Wrap Albuquerque Little Theatre's 87th Season'': \textit{The ALT production of Mary Poppins runs until June 12th, with performances on Fridays, Saturdays, and Sundays. Tickets are available through the ALT website or at the box office. Don't miss your chance to experience this magical show and be a part of the ALT's 87th season.}

``Rutherford Co. deputies searching for missing woman - FOX Carolina 21'': \textit{The disappearance of Patricia Murray has sent shockwaves through the community, and the Rutherford County Sheriff's Office is committed to doing everything they can to bring her home safely. As the search continues, the community remains hopeful that she will be found soon, and that she will be reunited with her loved ones.}

``Southfields Residential Care: A Comprehensive Guide'': \textit{In conclusion, at Southfields Residential Care, we understand the importance of providing a safe, supportive, and welcoming environment for our residents. We strive to create a place where people can live comfortably and safely, while still maintaining their independence and individuality. If you're looking for a care home that truly cares, look no further than Southfields Residential Care.}


\section{Comparison of original and regenerated CCNews}%

We show a direct comparison of human-generated, \llama-V2-generated, and \llama-V3-generated CCNews data in \autoref{tab:ccnewsexamples}. Anecdotally, the human-generations appear less similar to the \llama\ ones than the \llama\ ones appear to each other.
\begin{table*}[ht!]
    \centering
    \small
    \begin{tabular}{p{0.09\linewidth}p{0.2\linewidth}p{0.62\linewidth}}
    \toprule
Dataset & Prompt Title & Text \\ \midrule
Human & ARKit 1.5 Demos Showing Up Online & \textit{iOS 11.3 developer beta has been out for a few days and we're seeing some ARKit 1.5 examples showing up on YouTube and Twitter. The demos are rudimentary right now, but will likely improve as developers get more time with Apple's updates augmented reality technology. One of the big new features in ARKit 1.5 is vertical plane, or wall, detection. That's what the early demos are focusing on.}\\

Llama-V2   & ARKit 1.5 Demos Showing Up Online & \textit{\textcolor{red}{With its i}mproved wall \textcolor{red}{detection}, lighting estimation, and other features, \textcolor{red}{ARKit 1.5 is set to revolutionize} the way we interact with \textcolor{red}{the world} around us. As \textcolor{red}{developers} continue to explore the \textcolor{red}{potential} of this technology, we can expect to see even more innovative and practical applications of AR in the months and years to come. Whether you're a tech enthusiast, a developer, or simply someone interested in the future of technology, ARKit 1.5 demos are certainly worth keeping an eye on.}\\

Llama-V3  & ARKit 1.5 Demos Showing Up Online & \textit{\textcolor{red}{With its i}ntroduction of vertical plane \textcolor{red}{detection, ARKit 1.5 is set to revolutionize the world} of augmented reality, enabling \textcolor{red}{developers} to create more immersive, interactive, and engaging experiences. As the technology continues to evolve, it will be exciting to see how developers and companies harness its \textcolor{red}{potential} to transform industries and revolutionize the way we interact with the world around us. With the release of iOS 11.3 and ARKit 1.5, Apple is once again demonstrating its commitment to innovation and its desire to push the boundaries of what is possible with \textcolor{red}{technology}.} \\ \midrule

Human & Magical Show Mary Poppins to Wrap Albuquerque Little Theatre's 87th Season & \textit{The jack-of-all trades, Bert, introduces us to England in 1910 and the troubled Banks family. Young Jane and Michael have sent many a nanny packing before Mary Poppins arrives on their doorstep. Start the conversation, or Read more at BroadwayWorld.com.}\\
Llama-V2 & Magical Show Mary Poppins to Wrap Albuquerque Little Theatre's 87th Season & \textit{\textcolor{red}{Mary Poppins, the beloved musical} that \textcolor{orange}{the hearts of audiences for decades}, \textcolor{red}{is set to} wrap up \textcolor{red}{the 87th season of the Albuquerque Little Theatre (ALT) in New Mexico}. The show, which opened on May 13th, \textcolor{orange}{has been delighting audiences} with its magical blend of song, dance, and storytelling. \textcolor{red}{Based on the} popular Disney film of the same name, Mary Poppins \textcolor{red}{tells the} \textcolor{orange}{story} \textcolor{red}{of the Banks family, who are} in search of a new nanny to care for their children, Jane and Michael. The family's fortunes take a turn for the better when Mary Poppins, a practically perfect nanny, arrives on their doorstep. With her enchanting voice, graceful movements, and a sprinkle of magic, Mary wins over the hearts of the Banks family and their friends, \textcolor{orange}{bringing joy and order to their lives}}\\

Llama-V3 & Magical Show Mary Poppins to Wrap Albuquerque Little Theatre's 87th Season & \textit{\textcolor{red}{The beloved musical, Mary Poppins, is set to} conclude \textcolor{red}{the Albuquerque Little Theatre's (ALT) 87th season}, bringing a spoonful of sugar and a dash of magic to the city's theater scene. The show, directed by Henry Avery, promises to be a spectacular finale to a season that has seen a wide range of productions, from classic dramas to contemporary comedies. The story of Mary Poppins \textcolor{orange}{has been enchanting audiences for generations}, and the ALT production is expected to be no exception. The musical, \textcolor{red}{based on the} books by P.L. Travers, \textcolor{red}{tells the} \textcolor{orange}{tale} \textcolor{red}{of the Banks family, who are} struggling to \textcolor{orange}{find joy and discipline in their lives}. The arrival of Mary Poppins, played by actress Katie Calahan, \textcolor{orange}{brings a much-needed injection of fun and discipline into the household}, as she teaches the children valuable lessons about the importance of imagination, hard work, and family. }\\
\bottomrule
    \end{tabular}
    \caption{Examples of how \llama-regenerated CCNews text differs from the original. The original is presented in its entirety, the regenerate examples came from the end of the article in the case of ARKit 1.5, and from the middle in the case of Mary Poppins. \textcolor{red}{Red} denotes a near exact text match between \llama-V2 and \llama-V2 CCNews, \textcolor{orange}{Orange} denotes less exact match, potentially with position shifts across the sentence. }
    \label{tab:ccnewsexamples}
\end{table*}


\end{document}